\documentclass{article} 
\usepackage{iclr2016_conference,times}
\usepackage{hyperref}
\usepackage{url}

\usepackage{framed}
\usepackage{algorithmic}
\usepackage{graphicx}
\usepackage{amssymb}
\usepackage{amsmath}
\usepackage{amsthm}
\usepackage{caption}
\usepackage{subcaption}
\usepackage{listings}

\usepackage{verbatimbox}
\usepackage{adjustbox}
\usepackage{array}

\usepackage{tabularx}
\usepackage{makecell}
\usepackage{diagbox}
\usepackage{multicol}
\usepackage{multirow}

\newcolumntype{R}[2]{%
    >{\adjustbox{angle=#1,lap=\width-(#2)}\bgroup}%
    l%
    <{\egroup}%
}
\newcommand*\rot{\multicolumn{1}{R{45}{1em}}}

\usepackage[compact]{titlesec}
\titlespacing{\section}{0pt}{0.5ex}{0.3ex}
\titlespacing{\subsection}{0pt}{0.2ex}{0ex}
\titlespacing{\subsubsection}{0pt}{0.1ex}{0ex}

\usepackage{paralist}

\makeatletter
\ifcase \@ptsize \relax
  \newcommand{\miniscule}{\@setfontsize\miniscule{4}{5}}
\or
  \newcommand{\miniscule}{\@setfontsize\miniscule{5}{6}}
\or
  \newcommand{\miniscule}{\@setfontsize\miniscule{5}{6}}
\fi
\makeatother

\newcommand {\aplt} {\ {\raise-.5ex\hbox{$\buildrel<\over\sim$}}\ }

\newcommand{\fig}[1]{Fig.~\ref{fig:#1}}
\newcommand{\tab}[1]{Table~\ref{tab:#1}}
\newcommand{\secc}[1]{Section~\ref{sec:#1}}

\usepackage[symbol*]{footmisc}

\DefineFNsymbolsTM{myfnsymbols}{
  \textasteriskcentered *
  \textdagger    \dagger
  \textdaggerdbl \ddagger
  \textsection   \mathsection
  \textbardbl    \|%
  \textparagraph \mathparagraph
}%

\let\oldcite\cite
\renewcommand{\cite}[1]{[\oldcite{#1}]}

\title{MazeBase: \\A Sandbox for Learning from Games}

\author{Sainbayar Sukhbaatar\\
Department of Computer Science\\
Courant Institute\\
New York University \\
\texttt{sainbar@cs.nyu.edu} \\
\AND
Arthur Szlam, Gabriel Synnaeve, Soumith Chintala \& Rob Fergus \\
Facebook AI Research \\
New York \\
\texttt{\{aszlam,gab,soumith,robfergus\}@fb.com}
}

\begin{document}

\maketitle
\vspace{-5mm}
\begin{abstract}
  This paper introduces MazeBase: an environment for simple 2D  games,
  designed as a sandbox for machine learning approaches to reasoning
  and planning. Within it, we create 10 simple games embodying a range
  of algorithmic tasks (e.g. if-then statements or set negation). A
  variety of neural models (fully connected, convolutional network,
  memory network) are deployed via reinforcement learning on these
  games, with and without a procedurally generated
  curriculum. Despite the tasks' simplicity, the performance of the
  models is far from optimal, suggesting directions for future
  development. We also demonstrate the versatility of MazeBase by
  using it to emulate small combat scenarios from StarCraft. Models trained
  on the MazeBase version can be directly applied to StarCraft, where
  they consistently beat the in-game AI.
\end{abstract}

\vspace{-5mm}
\section{Introduction}

Games have had an important role in artificial intelligence research
since the inception of the field. Core problems such as search and
planning can explored naturally in the context of chess
or Go \citep{bouzy2001computer}. More recently, they have served as a test-bed for
machine learning approaches \citep{gvgai}. For example, Atari
games \citep{bellemare13arcade} have been investigated using neural models with reinforcement
learning \citep{mnih-atari-2013,NIPS2014_5421,mnih-dqn-2015}. The GVG-AI competition \citep{gvgai} uses a suite of 2D
arcade games to compare planning and search methods.

In this paper we introduce the MazeBase game environment, which
complements existing frameworks in several important ways.
\begin{itemize}
\item The
emphasis is on learning to understand the environment, rather than on testing algorithms for
search and planning. The framework deliberately lacks any simulation facility, and hence agents cannot use search methods to determine
the next action unless they can predict future game states themselves.  On the other hand, game components are meant to be reused in 
different games, giving models the opportunity to comprehend the
function of, say, a water tile. Nor are rules of the games provided to
the agent, instead they must be learned through exploration of the
environment. 
\item 
The environment has been designed to allow programmatic control
over the game difficulty. This allows the automatic contruction of
curricula \citep{Bengio09}, which we show to be important for training complex
models.  
\item Our games are based around simple algorithmic
reasoning, providing a natural path for exploring complex  abstract
reasoning problems in a language-based grounded setting. This contrasts with most
games that were originally designed for human enjoyment, rather than
any specific task. It also differs from the recent surge of work on
learning simple algorithms \citep{NTM_reinforce,Graves14,Zaremba15},
which lack grounding.
\item  Despite the 2D nature of the
environment, we prefer to use a text-based, rather than pixel-based,
representation.  This provides an efficient but expressive
representation without the
overhead of solving the perception problem inherent in pixel based representations.   
It easily allows for different task specifications and easy generalization of the models to other game
settings. We demonstrate this by training models in MazeBase and then
successfully evaluating them on
StarCraft\texttrademark\footnote{StarCraft and Brood War are
  registered trademarks of Blizzard Entertainment, Inc.}.
  See \citet{Mikolov15} for further discussion.
\end{itemize}

Using the environment, we introduce a set of 10 simple games and use
them to train range of standard neural network-based models (MLPs and
Convnets) via policy gradient \citep{Williams92simplestatistical}. We also combine the recent
MemN2N model \citep{end2endmemnn} with a reinforcement learning and evaluate it on
the games. The results show that current approaches struggle, despite
the relatively simple nature of tasks. They also  highlight clear areas for future model
exploration, but we defer this for further work. MazeBase is
an open-source platform, implemented using Torch and can be downloaded
from \url{https://github.com/facebook/MazeBase}.

\subsection{Related Work}

The MazeBase environment can be thought of as a small practical step
towards of some of the ideas discussed at length in \citet{Mikolov15}.
In particular, interfacing the agent and the environment in
(quasi-)natural language was inspired by discussions with the authors
of that work.  However, our ambitions are more local, focusing on
the border where current models fail (but nearly succeed), rather than
aiming for a global view of a path towards AI.  For example, we
specifically avoid algorithmic tasks that require unbounded recursions
or loops, as we find that there is plenty of difficulty in learning
simple if-then statements.  Furthermore, for the example games
described below, we allow large numbers of training runs, as the noise
from reinforcement with discrete actions remains challenging even
with many
samples.  

In non-game environments, there has been recent work on learning
simple algorithms. \citep{Graves14,pointer_networks,Joulin15,
  NTM_reinforce} demonstrating tasks such as sorting and reversal of
inputs. The algorithms instantiated in our games are even simpler,
e.g. conditional statements or navigation to a location, but involve
interaction with an environment. In some of these approaches
\citep{mnih-atari-2013,NIPS2014_5421,mnih-dqn-2015,Joulin15,
  NTM_reinforce} the models were trained with reinforcement learning
or using discrete search, allowing possibly delayed rewards with
discrete action spaces. Our games also involve discrete actions, and
these works inform our choice of the reinforcement learning
techniques.
Several works have also demonstrated the ability of neural models to learn to
answer questions in simple natural language within a restricted environment
\citep{Weston14,end2endmemnn}.  The tasks we present here share many features
with those in \citet{Weston15}, and the input-output format our games
use is inter-operable with their stories.  However,
 during training and testing, the environment in \citet{Weston15} is static,
 unlike the game worlds we consider.  

 Developing AI for game agents has an extensive literature.  Our work
 is similar to \citet{mnih-atari-2013,NIPS2014_5421,mnih-dqn-2015} in
 that we use reinforcement and neural models when training on
 games.  The GVG-AI competition \citep{gvgai} is similar in overall
 intent to MazeBase, but differs in that it is more appropriate for
 testing search-based methods, since a simulator (and game rules) are
 provided. Correspondingly, many of the top algorithms in the
 competition rely on Monte-Carlo tree search methods. In contrast,
 MazeBase is designed to be a sandbox that supports the development of learning-based algorithms;
 any search done by an agent must be done with the agent's own predictions.  
  Similarly, 
competitions have been organized around Super Mario \citep{Shaker10}, and Pacman \citep{Ikehata11}
and encourage search based on heuristics of the specific game.   On the other hand, MazeBase is designed 
to encourage learning algorithms that can understand the environment and reuse knowledge
between games, and to be used for the incremental exploration of
 certain core AI problems, for example the basic logical reasoning addressed in this
 paper.  Furthermore \citep{Shaker10}, \citep{Ikehata11} do not easily support
an algorithmic curriculum for training.  
  Also note that given the sandbox nature of MazeBase, in principle it could be
 used to recreate any of these games, including those in the ACE
 benchmark \citep{bellemare13arcade}. The versatility of MazeBase is
 demonstrated by the ease with which we were able to create a proxy
 for StarCraft combat within the environment.   Our environment takes many basic ideas from the classical 
 Puddle World \cite{Rajendran15}, 
 for example the basic 2-$d$ grid structure and water obstacles, but is richer, and the agent is not expected
 to memorize any given world, as they are regenerated at each new game, and agents are tested on unseen worlds.

\section{Environment and tasks}
Each game is played in a 2D rectangular grid.  In the specific examples below, the dimensions range
from 3 to 10 on each side, but of course these can be set however the user likes.  Each location in the grid can be empty, or may contain one
or more items.  The agent can move in each of the four cardinal
directions, assuming no item blocks the agents path.  The items in the
game are:
\begin{itemize}
\item 
{\bf Block:}  an impassible obstacle that does not allow the agent to move to
that grid location.
\item 
{\bf Water:}  the agent may move to a grid location with water, but incurs
an additional cost of (fixed at $-0.2$ in the games below) for doing so.
\item 
{\bf Switch:}  a switch can be in one of $m$ states, which we refer to as
colors.  The agent can toggle through the states cyclically by a toggle action
when it is at the location of the switch.
\item 
{\bf Door:}  a door has a color, matched to a particular switch.  The agent
may only move to the door's grid location if the state of the switch
matches the state of the door.
\item 
{\bf PushableBlock:}  This block is impassable, but can be moved with a separate ``push'' actions.  
The block moves in the direction of the push, and the agent must be located adjacent to 
the block opposite the direction of the push.
\item 
{\bf Corner:}  This item simply marks a corner of the board.
\item
{\bf Goal:}  depending on the task, one or more goals may exist, each
  named individually.
\item 
{\bf Info:} these items do not have a grid location, but can specify a
task or give information necessary for its completion.
\end{itemize}

The environment is presented to the agent as a list of sentences, 
each describing an item in the game. For example, an agent might see
``Block at [-1,4]. Switch at [+3,0] with blue color. Info: change switch to red.''
Such representation is compatible with the format of the bAbI tasks, introduced in \citet{Weston15}.
However, note that we use egocentric spatial coordinates (e.g.~ the goal G1 in Fig.~1 (left) is at coordinates [+2,0]), meaning that
the environment updates the locations of each object after an
action\footnote{This is consistent with \citet{end2endmemnn}, where the ``agent''
answering the questions was also given them in egocentric temporal
coordinates.}. 
Furthermore, for tasks involving multiple goals, we have two versions of the game.  
In one, the environment automatically sets a flag on visited goals. 
In the harder versions, this mechanism is absent but the agent has a special action that allows it to release a ``breadcrumb'' into the environment, enabling it to record locations it has visited.   
In the experiments below, unless otherwise specified, we report results on games with the explicit flag.

The environments are generated randomly with some distribution on the
various items.  For example, we usually specify a uniform distribution
over height and width (between 5 and 10 for the experiments reported here), and a percentage of wall blocks and
water blocks (each range randomly from 0 to 20\%).
\begin{figure}[h!]
\centering
\mbox{
\fbox{\includegraphics[width=0.4\linewidth]{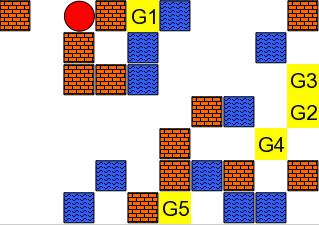}}\hspace{3mm}
\fbox{\includegraphics[width=0.4\linewidth]{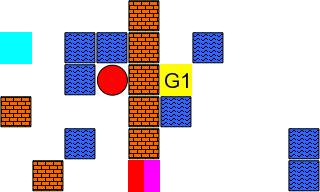}}
}
\caption{Examples of Multigoal (left) and Light Key (right)
tasks. Note  that the layout and dimensions of the environment varies
between different  instances of each task (i.e.~the location and quantity
of walls,  water and goals all change). The  agent is shown as a red blob
and the goals are shown in yellow. For  LightKey, the switch is show in
cyan and the door in magenta/red  (toggling the switch will change the
door's color, allowing it to  pass through).}
\label{fig:examples}
\end{figure}

\subsection{Tasks}

Although our game world is simple, it allows for a rich variety of
tasks. In this work, we explore those that require different
algorithmic components first in isolation and then in
combination. These components include: 
\begin{itemize}
\item {\bf Set operations:} iterating through a list of goals, or
  negation of a list, i.e. all items {\em except} those specified in a list.
\item {\bf Conditional reasoning:}  {\em if-then} statements, {\em while}
  statements.  
\item {\bf Basic Arithmetic:} comparison of two small numbers.
\item {\bf Manipulation:} altering the environment by toggling a {\bf Switch}, or moving a {\bf PushableBlock}.
\end{itemize}
These were selected as key elements needed for complex reasoning
tasks, although we limit ourselves here to only combining a them few
of them in any given task. We note that most of these have direct
parallels to the bAbI tasks (except for {\bf
  Manipulation} which is only possible in our non-static
environment). We avoid tasks that require
unbounded loops or recursion, as in \citet{Joulin15}, and instead view ``algorithms''
more in the vein of following a recipe from a cookbook.   In particular, we want our agent
to be able to follow directions; the same game world may host multiple
tasks, and the agent must decide what to do based on the ``Info''
items.  As we demonstrate, standard neural models find this to be
challenging. 

In all of the tasks, the agent incurs a fixed penalty for each action
it makes, this encourages the agent to finish the task promptly.  In
the experiments below, this is set to 0.1. In addition, stepping on a
Water block incurs an additional penalty of 0.2. For most games, a
maximum of 50 actions are allowed. The tasks define extra penalties
and conditions for the game to end.

\begin{itemize}
\item 
{\bf Multigoals:} In this task, the agent is given an ordered
list of goals as ``Info'', and needs to visit the goals in that order.
In the experiments below, the number of goals ranges from 2 to 6, and the number of
``active'' that the agent is required to visit ranges from 1 to 3 goals.  The
agent is not given any extra penalty for visiting a goal out of order, but
visiting a goal before its turn does not count towards visiting all goals.
The game ends when all goals are visited. This task involves the
algorithmic component of iterating over a list.
\item
{\bf Exclusion:} The ``Info'' in this game specifies a list of
goals to avoid.  The agent should visit all other unmentioned goals. The
number of all goals ranges form 2 to 6, but the number of active goals
ranges from 1 to 3. As in the Conditional goals game, the
agent incurs a 0.5 penalty when it steps on a forbidden goal. This
task combines {\bf Multigoals} (iterate over set) with set negation.
\item
{\bf Conditional Goals:} In this task, the destination goal is
conditional on the state of a switch.  The ``Info'' is of the form ``go to
goal $g_i$ if the switch is colored $c_j$, else go to goal $g_l$.''  In
the experiments below, the number of the number of colors range from 2 to 6 and
the number of goals from 2 to 6.  Note that there can be more colors than
goals or more goals than colors.  The task concludes when the agent
reaches the specified goal; in addition, the agent incurs a 0.2 penalty for
stepping on an incorrect goal, in order to encourage it to read the
info (and not just visit all goals). The task requires conditional
reasoning in the form of an {\em if-then} statement. 
\item
{\bf Switches:} In this task, the game has a random number of
switches on the board.  
The agent is told via the ``Info'' to toggle
all switches to the same color, and the agent has the choice of color; to
get the best reward, the agent needs to solve a (very small) traveling
salesman problem.  
 In the experiments below,
the number of switches ranges from 1 to 5 and the number of colors from 1
to 6.  The task finishes when the switches are correctly toggled.  There
are no special penalties in this task.  The task instantiates a form
of {\em while} statement. 
\item 
{\bf Light Key:} In this game, there is a switch and a door in a
wall of blocks.   The agent should navigate to a goal which may be on the
wrong side of a wall of blocks.  If the goal is on the same side of the
wall as the agent, it should go directly there; otherwise, it needs move
to and toggle the switch to open the door before going to the goal.  There
are no special penalties in this game, and the game ends when the agent
reaches the goal. This task combines {\em if-then} reasoning with
environment manipulation.
\item
{\bf Goto:} In this task, the agent is given an absolute location on the grid as a target.
The game ends when the agent visits this location.   Solving this task requires the agent to 
convert from its own egocentric coordinate representation to absolute
coordinates. This involves comparison of small numbers.
\item 
{\bf Goto Hidden:} In this task, the agent is given a list of goals with absolute coordinates, and then
is told to go to one of the goals by the goal's name.   The agent is not directly given the goal's location, it must
read this from the list of goal locations. The number of goals ranges
from 1 to 6. The task also involves very simple comparison operation.
 \item
{\bf Push block:} In this game, the agent needs to push a Pushable block so that it lays on top
of a switch. Considering the large number of actions needed to solve this task, 
the map size is limited between 3 and 7, and the maximum block and
water percentage is reduced to 10\%. The task requires manipulation of
the environment.  
 \item
{\bf Push block cardinal:} In this game, the agent needs to push a
Pushable block so that it is on a specified edge of the maze, e.g. the
left edge. Any location along the edge is acceptable. 
The same limitation as Push Block game is applied. 
\item
{\bf Blocked door:} In this task, the agent should navigate to a goal which may lie on the opposite
side of a wall of blocks, as in the Light Key game.  However, a PushableBlock blocks the gap in the wall instead of 
a door. This requires {\em if-then} reasoning, as well as environment manipulation.
 
\end{itemize}

For each task, we compute offline the optimal solution. For some of
the tasks, e.g. {\bf Multigoals}, this involves solving a traveling
salesman problem (which for simplicity is done approximately). This provides
an upper bound for the reward achievable. This is used for comparison
purposes only, i.e.~it is not used for training the models.

With the exception of the ~{\bf Multigoals} task, all these are Markovian; and ~{\bf Multigoals} is Markovian with 
the explicit ``visited'' flag, which we use in the experiments below.  Nevertheless, the tasks are not at all simple; although the 
environment can easily be used to build non-Markovian tasks, we find that the solving these tasks without the agent having to reason about its past actions
is already challenging. Examples of each game are shown at \url{https://youtu.be/kwnp8jFRi5E}. 
Note that building new tasks is an easy operation in the MazeBase environment, indeed many of those above are implemented in few hundred lines of code.



\section{Models}
We investigate several different types of model: (i) simple linear,
(ii) multi-layer neural nets,
(iii) convolutional nets and (iv) end-to-end memory networks
\citep{Weston14,end2endmemnn}. While the input format is quite
different for each approach (detailed below), the outputs are the
same: a probability distribution over set of discrete actions \{N,S,E,W,toggle
switch,push-N,push-S,push-E,push-W\}; and a continuous baseline value 
predicting the expected reward.
We do not consider models that are recurrent in the state-action
sequence such as RNNs or LSTMs, because as discussed above, these
tasks are Markovian.

\noindent {\bf Linear:} For a simple baseline we take the existence of
each possible word-location pair on the largest grid we consider
($10\times 10$) and each ``Info'' item as a separate feature, and
train a linear classifier to the action space from these features.
To construct the input, we take bag-of-words (excluding location words) representation of all items at the same location.
Then, we concatanate all those features from the every possible locations and info items. 
For example, if we had $n$ different words and $w \times h$ possible locations with $k$ additional info items, then the input dimension would be $(w \times h + k) \times n$.

\noindent {\bf Multi-layer Net:} 
Neural network with multiple fully connected layers separated by tanh non-linearity. 
The input representation is the same as the linear model.

\noindent {\bf Convolutional Net:} 
First, we represent each location by bag-of-words in the same way as linear model.
Hence the environment is presented as a 3D cube of size $w \times h \times n$, 
which is then feed to four layers of convolution 
(the first layer has $1 \times 1$ kernel, which essentially makes it an embedding of words).  
Items without spatial location (e.g. ``Info'' items) are each represented as a bag of words,
and then combined via a fully connected
layer to the outputs of the convolutional layers; these are then
passed through two fully connected layers to output the actions (and
a baseline for reinforcement).


\noindent {\bf Memory Network:} Each item in the game (both physical
items as well as ``info'') is represented as bag-of-words
vectors. The spatial location of each item is also represented as a
word within the bag. E.g.~ a red door at [+3,-2] becomes the vector
\{red\_door\} + \{x=+3,y=-2\}, where \{red\_door\} and \{x=+3,y=-2\}
are word vectors of dimension 50. These embedding vectors will be
learned at training time. As a consequence, the memory network has to
learn the spatial arrangement of the grid, unlike the convolutional
network.  Otherwise, we use the architecture from \citep{end2endmemnn}
with 3 hops and tanh nonlinearities.


\section{Training Procedures}





We use policy gradient \citep{Williams92simplestatistical} for training,
which maximizes the expected reward using its unbiased gradient estimates.  
First, we play the game by feeding the current state $x_t$ to the model,
and sampling next action $a_t$ from its output. 
After finishing the game, we update the model parameters $\theta$ by
\[
\Delta \theta = \sum_{t=1}^T \left[ 
  \frac{\partial \log p(a_t| x_t, \theta)}{\partial \theta} 
  \left(\sum_{i=t}^T r_i - b \right) 
\right],
\]
where $r_t$ is reward given at time $t$, and $T$ is the length of the game.

Instead of using a single baseline $b$ value for every state,
we let the model output a baseline value specific to the current state.
This is accomplished by adding an extra head to models for outputting the baseline value.
Beside maximizing the expected reward with policy gradient, the
models are also trained to minimize the distance between the baseline value and actual reward.   
The final update rule is
\[
\Delta \theta = \sum_{t=1}^T \left[ 
  \frac{\partial \log p(a_t| x_t, \theta)}{\partial \theta} 
  \left(\sum_{i=t}^T r_i - b(x_t, \theta) \right) 
   - \alpha
  \frac{\partial}{\partial \theta} 
  \left(\sum_{i=t}^T r_i - b(x_t, \theta) \right)^2 
\right].
\]
Here hyperparameter $\alpha$ is for balancing the two objectives, which is set to 0.03 in all experiments.
The actual parameter update is done by RMSProp \citep{Tieleman2012} with
learning rates optimized for each model type.


For better parallelism, the model plays and learns from 512 games simultaneously, which spread on multiple CPU threads.
Training is continued for 20 thousand such parallel episodes, which amounts to 10M game plays.
Depending on the model type, the whole training process took from few hours to few day on 18 CPUs of a single machine.
\subsection{Curriculum}
\begin{figure}[h!]
\centering
\mbox{
\mbox{\includegraphics[width=0.33\linewidth]{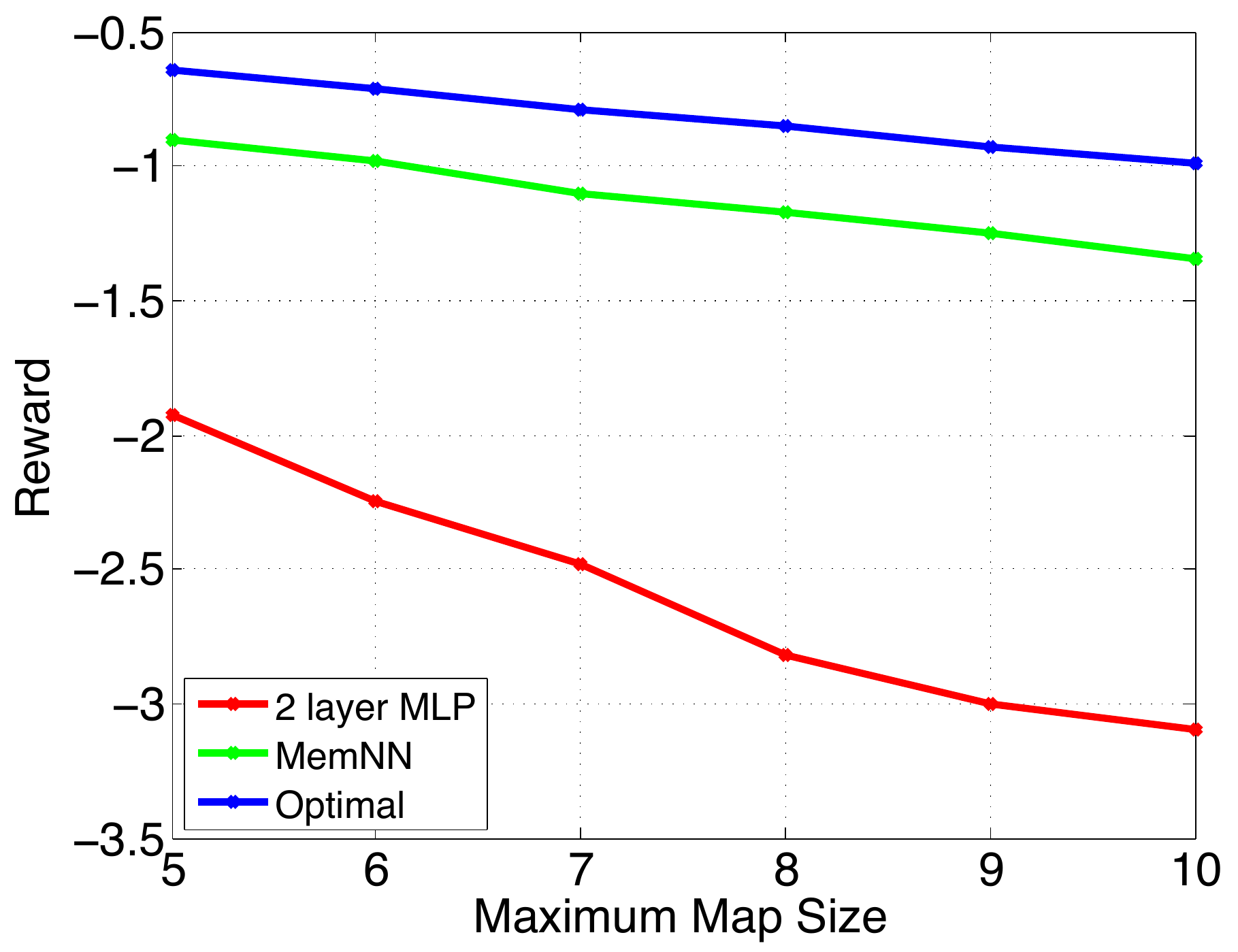}}
\mbox{\includegraphics[width=0.33\linewidth]{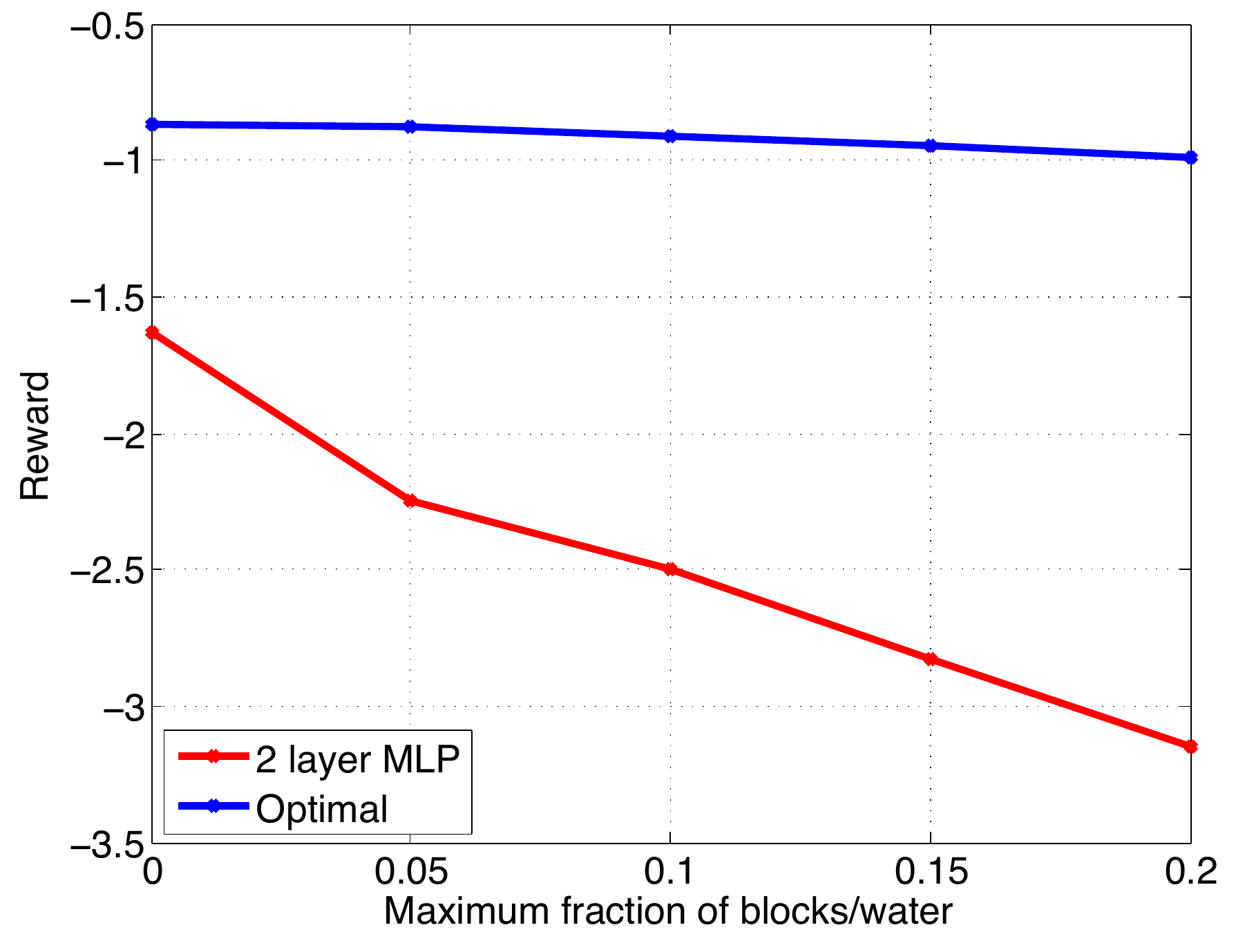}}
\mbox{\includegraphics[width=0.33\linewidth]{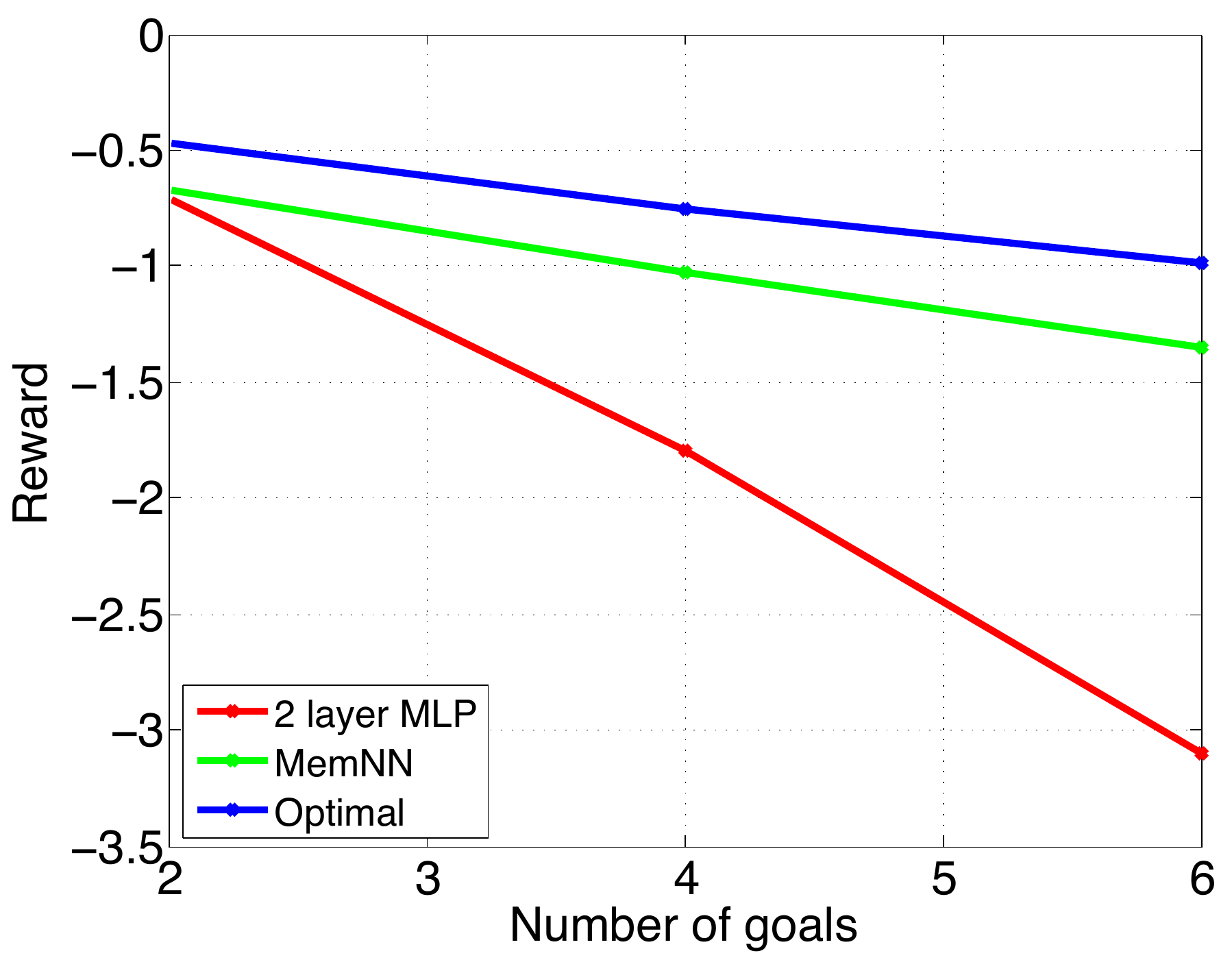}}
}
\caption{In the MazeBase environment, the difficulty of tasks can be
  varied programmatically. For example, in the {\bf Multigoals} game the maximum map size, fraction
  of blocks/water and number of goals can all be varied. This affects
  the  difficulty of tasks, as shown by the optimal reward (blue
  line). It also reveals how robust a given model is to task
  difficulty. For a 2-layer MLP (red line), the reward achieved
  degrades much faster than the MemNN modle (green line) and the inherent task difficulty.}
\label{fig:curriculum}
\end{figure}

A key feature of our environment is the ability to
programmatically vary all the properties of a given game. We use this
ability to construct instances of each game whose difficulty is
precisely specified (see \fig{curriculum}). These instances can then be shaped into a
curriculum for training \citep{Bengio09}. As we demonstrate, this is very important for
avoiding local minima and helps to learn superior models. 

Each game has many variables that impact the difficulty. Generic ones
include: maze dimensions (height/width) and the fraction of blocks \&
water. For switch-based games ({\bf Switches}, {\bf Light Key})
the number of switches and colors can be varied. For goal based games
({\bf Multigoals},{\bf Conditional Goals}, {\bf Exclusion}, the
variables are the number of goals (and active goals). For the combat
game {\bf Kiting} (see \secc{combat}), we vary the number of agents \&
enemies, as well as their speed and their initial health.  

The curriculum is specified by an upper and lower success thresholds
$T_u$ and $T_l$ respectively. If the success rate of the model falls
outside the $[T_l,T_u]$ interval, then the difficulty of the generated
games is adjusted accordingly. Each game is generated by uniformly
sampling each variable that affects difficulty over some range. The
upper limit of this range is adjusted, depending on which of $T_l$ or
$T_u$ is violated. Note that the lower limit remains unaltered, thus
the easiest game remains at the same difficulty. For the last third of
training, we expose the model to the full range of difficulties by
setting the upper limit to its maximum preset value. 



\section{Experimental Results}

\begin{figure}[h!]
\begin{center}
\includegraphics[width=\linewidth]{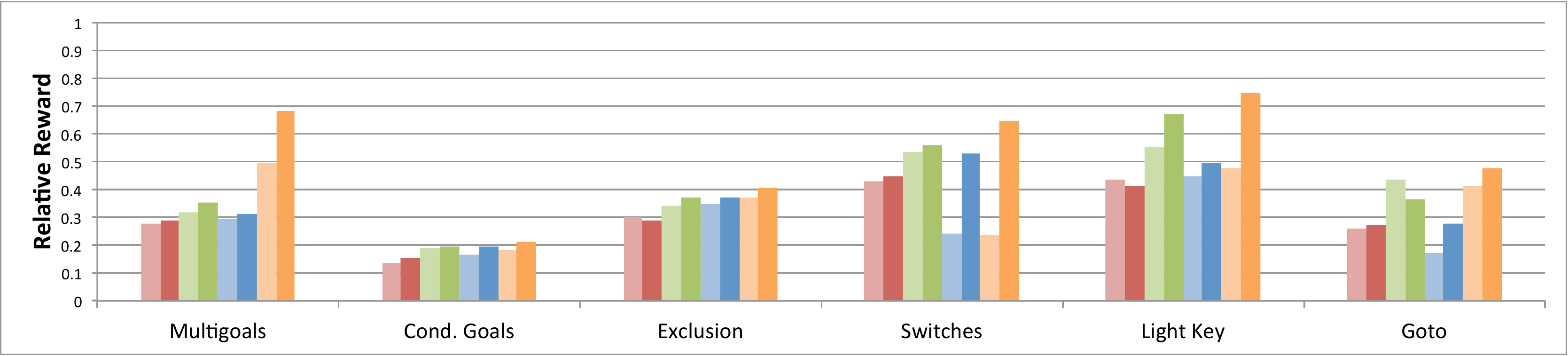}
\includegraphics[width=\linewidth]{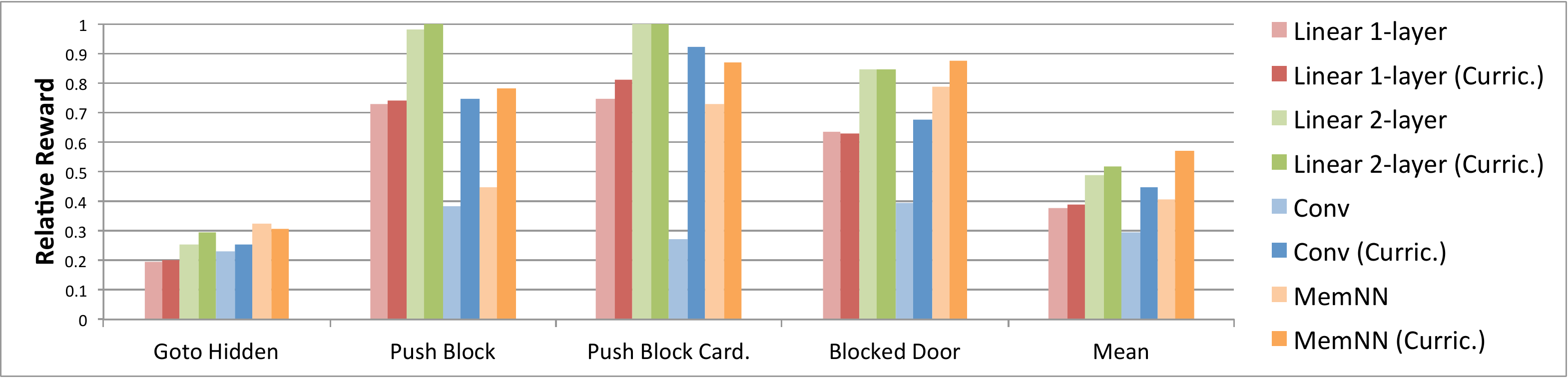}
\end{center}
\caption{Reward for each model jointly trained on the 10 games, with
  and without the use of a curriculum during training. The $y$-axis shows
  relative reward (estimated optimal reward / absolute reward), thus higher is
  better. The estimated optimal policy corresponds to a value of 1 and a value
  of 0.5 implies that the agent takes twice as many steps to complete
  the task as is needed (since most of the reward signal comes from
  the negative increment at each time step).}
\label{fig:bar}
\end{figure}

\fig{bar} shows the performance of different models on the games. Each model is trained
{\it jointly} on all 10 games. Given the stochastic nature of
reinforcement learning, we trained each model 10 times and picked the
single instance that had the highest mean reward across all
  tasks (i.e.~the same model is evaluated on all 10 tasks). \tab{result} in the Appendix gives the max,mean and standard
devision of rewards for each task and method. A video showing a
trained MemNN model playing each of the games can be found at \url{https://youtu.be/kwnp8jFRi5E}. 
The results revealed a number of interesting points.

On many of the games at least some of the models were able to learn a
  reasonable strategy.  The models were all able to learn to convert between egocentric and absolute coordinates by using the corner blocks.  They could respond appropriately to the different arrangements of the {\bf Light Key} game, and make decent decisions on whether to try to go directly to the goal, or to first open the door.  The 2-layer networks were able to completely solve the the tasks with pushable blocks.

That said, despite the simplicity of the games, and the number of trials allowed, the models were not able to completely solve
  (i.e.~discover optimal policy) most of the games:
  \begin{itemize}
\item On {\bf Conditional Goals} and {\bf Exclusion},  all
  models did poorly. On inspection, it appears they adopted the
  strategy of blindly visiting all goals, rather than visiting the
  correct one.     
\item With some of the models, we were able to train jointly, but make a few of the game types artificially small; then at test time
successfully run those games on a larger map.  The models were able to learn the notion of the locations independently from the task (for locations they had seen in training).  
On the other hand, we tried to test the models above on unseen tasks that were never shown at train time, but used the same vocabulary 
(for example: ``go to the left'', instead of ''push the block to the left'').  None of our models were able to succeed, highlighting how far we are from operating at a ``human level'', even in this extremely restricted setting.  
\end{itemize}
With respect to comparisons between the models: 
\begin{itemize}   
\item On average, the memory network did best out of the
  methods. However, on the games with pushable blocks,  the 2 layer neural nets were superior e.g.~{\bf Exclusion}
  and  {\bf Push Block} or the the same {\bf Blocked Door}.
  Although we also trained 3 layer neural net, the result are not included here
  because it was very similar to the rewards of 2 layer neural net.
\item The linear model did better than might be expected, and surprisingly,  the
  convolutional nets were the worst of the four models. However, the
  fully connected models had significantly more parameters than either the
  convolutional network or the memory network.  For example, the 2 layer neural net had a hidden layer
  of size 50, and a separate input for the outer product of each location and word combination.  Because of the size of 
  the games, this is 2 orders of magnitude more parameters than the convolutional network or memory network.  Nevertheless, even with 
  very large number of trials, this architecture could not learn many of the tasks.
\item The memory network seems superior on games involving decisions using information in the {\bf info} items
  (e.g.~{\bf Multigoals}) whereas the 2-layer neural net was better
  on the games with a pushable block (~{\bf Push
    Block},  ~{\bf Push
    Block Cardinal}, and {Blocked Door}).   Note that  because we use egocentric coordinates, for ~{\bf Push
    Block Cardinal}, and to a lesser extent ~{\bf Push
    Block}, the models can memorize all the local configurations of the block and agent.
\item All methods had a significant variance in performance
  over its 10 instances, except for the linear model. However, the curriculum
  significantly decreased the variance for all methods, especially for 2-layer neural net. 
  \end{itemize}

With respect to different training modalities:
  \begin{itemize}
\item The curriculum generally helped all approaches, but gave the
  biggest assistance to the memory networks, particularly for {\bf
    Push Block} and related games. 
\item We also tried supervised training (essentially imitation
  learning), but the results were more or less the same as for reinforcement.  The one exception was learning to use the Breadcrumb action 
  for the ~{\bf Multigoals} game.  None of our models were able to learn to use the breadcrumb to mark visited locations without supervision.
  Note that in the tables we show results with the explicit ``visited'' flag given by the environment.
\end{itemize}

\section{Combat Games}
\label{sec:combat}
In this section we use MazeBase to implement several simple
combat games. We train agents using these games and then test them on 
combat micro-management in StarCraft: Brood War, involving a limited number of
troops:
\begin{itemize}
    \item {\bf Kiting} (Terran Vulture vs Protoss Zealot): a match-up
        where we control a weakly armored fast ranged unit, against a more
        powerful but melee ranged unit. To win, our unit needs to alternate
        fleeing the opponents and shooting at them when its weapon has cooled
        down.
    \item {\bf Kiting hard} (Terran Vulture vs 2 Protoss Zealots):
        same as above but with 2 enemies.
    \item {\bf 2 vs 2} (Terran Marines): a symmetric match-up where both teams
        have 2 ranged units.
\end{itemize}
Our goal here is not to asses the ability of the models to
generalize, but rather whether we can make the game in our
environment close to its counterpart in StarCraft, to show that the environment can be used for prototyping scenarios where
training on an actual game may be technically challenging.

In MazeBase, the kiting scenario consists of a standard maze where an
agent aims to kill up to two enemy bots.  After each shot, the agent
or enemy is prevented from firing again for a small time interval
(cooldown).  We introduce an imbalance by (i) allowing the agent to
shoot farther than the enemy bot(s) and (ii) giving the agent
significantly less health than the bot(s); and by allowing the enemy
bot(s) to shoot more frequently than the agent (shorter cooldown).
The agent has a shot range of 7 squares, and the bots have a shot
range of 4 squares.  The enemy bot(s) moves (on average) at .6 the
speed of the agent.  This is accomplished by rolling a ``fumble'' each
time the bot tries to move with probability .4.  The agent has health
chosen uniformly between 2 and 4, and the enemy(s) have health
uniformly distributed between 4 and 11.  The enemy can shoot every 2
turns, and the agent can shoot every 6 turns.  The enemy follows a
heuristic of attacking the agent when in range and its cooldown is 0,
and attempting to move towards the agent when it is closer than 10
squares away, and ignoring when farther than 10 squares.

The 2 vs. 2 scenario is modeled in MazeBase with two agents, each of
which have 3 health points, and two bots, which have hitpoints
randomly chosen from 3 or 4.  Agents and bots have a range of 6 and a
cooldown of 3.  The bots use a heuristic of attacking the closest
agent if they have not attacked an agent before, and continuing to
attack and follow that agent until it is killed.
 
In both the kiting and 2x2 scenarios, we randomly add noise to the
agents inputs to account for the fact that it will encounter new
vocabulary when playing StarCraft.  That is, 10\% of the time, the
numerical value of the enemies' or agents' health, cooldown, etc is
taken to be a random value. We train a MemNN model using the difference between
the armies hit points, and win or loss of the overall battle, as the
reward signals.

We also run the scenarios inside StarCraft: Brood War. We used
\citet{BWAPI} to connect the game to our Torch framework. We can
receive the game state and send orders, enabling us to do a
reinforcement learning loop.  We train a 2 layer neural network and
MemNN models using the same protocol. The features used are all
categorical (as for MazeBase) and represent the hit points (health),
the weapon cooldown, and the x and y positions of the unit. We used a
multi-resolution encoding (coarser going further from the unit we
control) of the position on $256 \times 256$ map to reduce the number
of parameters.  In case of the multiple units, each is controlled
independently. We take an action every 8 frames (the atomic time unit
@ 24 frames/sec). The architectures and hyper-parameter settings are
the same as used in the {\bf Kiting} game (except the multi-resolution
feature map). 

We find that the models are able to learn basic tactics such as focusing their
fire on weaker opponents to kill them first (thus reducing the total amount of
incoming damage over the game). This results in a win rate of 80\% over the
built-in StarCraft AI on {\bf 2 vs 2}, and nearly perfect results on
{\bf Kiting} (see \tab{scresult}).   The video
\url{https://youtu.be/Hn0SRa_Uark} shows example gameplay of our MemNN
model for the {\bf StarCraft Kiting hard} scenario.    

Finally, we test the models trained in our environment directly on
StarCraft.  We make no modifications to the models, and minimal
changes to the interface (we scale the health by a factor of 10, and
x,y and cooldown values by a factor of 4).  The success rate of the
models trained in the maze but tested in Starcraft is comparable to
training directly in StarCraft, showing that our environment can be
effectively used as a sandbox for exploring model architectures and
hyper-parameter optimization.
\begin{table}
\begin{center}
\scriptsize
\begin{tabular}{|l||c|c|c|}
\hline
& {\bf 2 vs 2} & {\bf Kiting} & {\bf Kiting hard} \\ \hline
Attack weakest & 85\% &  0\% & 0\% \\ \hline
MemNN          & 80\%  & 100\%  & 100 \% \\ \hline
MemNN (transfer) & 65\%  & 96\%  & 72\%  \\ \hline

\end{tabular}
\caption{Win rates against StarCraft built-in AI.  The 2nd row shows
  the hand-coded baseline strategy of always attacking the weakest enemy
  (and not fleeing during cooldown). The 3rd row shows 
  MemNN trained and tested on StarCraft. The last row shows a
  MemNN trained entirely inside MazeBase and tested on StarCraft
  with no modifications or fine tuning except scaling of the inputs.}
\label{tab:scresult}
\end{center}
\end{table}

\section{Discussion}

The MazeBase enivronment allows easy creation of games and precise
control over their behavior. This allowed us to quickly to devise a
set of 10 simple games embodying algorithmic components and evalaute
them using a 
range of neural models. The flexibility of the environment enabled
curricula to be created for each game which aided the training of the
models and resulted in superior test performance. Even with the aid of
a curriculum, in most cases the models fell short of optimal
performance. The memory networks were able to solve some tasks that the
fully-connected models and convnets could not, although overall the
performance was similar. This suggests that existing neural models lack some
fundamental abilities that are needed to solve algorithmic
reasoning. Potential candidates include: the ability to plan or forecast the outcome of
actions and a more sophisticated memory (it is notable that the MemNN
outperformed the others on tasks with involving large {\bf info}
items). 

We also showed how the MazeBase environment can be used to develop
model architectures that when trained on StarCraft can convincingly
beat the in-game AI in a range of simple combat settings. More
indirectly, we can also use the environment to build games that
approximate a task of interest, enabling the training of models that
will perform effectively on the target task, without having exposure
to it during training.

\section*{Appendix}

\begin{table}[h!]
\begin{center}
\scriptsize
\setlength{\tabcolsep}{5pt}
\begin{tabular}{cc||c|c|c|c|c|c|c|c|c|c|c|}
\multicolumn{1}{c}{}
&
\multicolumn{1}{c}{}
&
\rot{Multigoals} &
\rot{Cond. Goals} &
\rot{Exclusion} &
\rot{Switches} &
\rot{Light Key} &
\rot{Goto} &
\rot{Goto Hid.} &
\rot{Push Block} & 
\rot{Push Block Card.} & 
\rot{Blocked Door} &
\rot{Mean}
    \\  \Xhline{2\arrayrulewidth}
\parbox[t]{2mm}{\multirow{12}{*}{\rotatebox[origin=c]{90}{ No Curriculum}}} & 
\multirow{3}{*}{Linear}                         
& -3.59 & -3.54 & -2.76 & -1.66 & -1.94 & -1.82 & -2.39 & -2.50 & -1.64 & -1.66 & -2.35 \\ &
& -3.67 & -3.93 & -2.62 & -1.65 & -1.89 & -1.78 & -2.48 & -2.54 & -1.65 & -1.63 & -2.37 \\ &
& $\pm$0.06 & $\pm$0.13 & $\pm$0.06 & $\pm$0.03 & $\pm$0.04 & $\pm$0.02 & $\pm$0.02 & $\pm$0.02 & $\pm$0.02 & $\pm$0.03 & $\pm$0.01 \\ 
\cline{2-13} & 
\multirow{3}{*}{2 layer NN}        
& -3.14 & -2.61 & -2.42 & -1.32 & -1.54 & -1.07 & -1.83 & -1.86 & -1.21 & -1.25 & -1.82 \\ &
& -3.46 & -3.11 & -2.89 & -1.69 & -2.17 & -1.83 & -2.46 & -2.79 & -2.14 & -2.08 & -2.46 \\ &
& $\pm$0.79 & $\pm$0.99 & $\pm$1.11 & $\pm$0.68 & $\pm$1.43 & $\pm$1.63 & $\pm$1.31 & $\pm$1.25 & $\pm$1.47 & $\pm$1.49 & $\pm$1.20 \\ 
\cline{2-13} & 
\multirow{3}{*}{ConvNet}           
& -3.36 & -2.90 & -2.38 & -2.96 & -1.90 & -2.70 & -2.06 & -4.80 & -4.50 & -2.70 & -3.03 \\ &
& -4.35 & -4.17 & -3.97 & -2.98 & -3.78 & -4.18 & -3.85 & -4.95 & -4.86 & -4.09 & -4.12 \\ &
& $\pm$0.83 & $\pm$1.06 & $\pm$1.32 & $\pm$0.05 & $\pm$1.57 & $\pm$1.06 & $\pm$1.47 & $\pm$0.05 & $\pm$0.18 & $\pm$1.16 & $\pm$0.85 \\ 
\cline{2-13}& 
\multirow{3}{*}{MemNN}             
& -2.02 & -2.70 & -2.22 & -2.97 & -1.78 & -1.14 & -1.44 & -4.06 & -1.68 & -1.34 & -2.19 \\ & 
& -3.68 & -3.51 & -3.06 & -2.98 & -2.72 & -2.25 & -3.42 & -4.47 & -2.71 & -2.43 & -3.13 \\ & 
& $\pm$0.99 & $\pm$1.04 & $\pm$1.33 & $\pm$0.03 & $\pm$1.56 & $\pm$1.89 & $\pm$1.19 & $\pm$0.68 & $\pm$1.54 & $\pm$1.77 & $\pm$1.14 \\ 
\Xhline{2\arrayrulewidth}
\parbox[t]{2mm}{\multirow{12}{*}{\rotatebox[origin=c]{90}{ Curriculum}}} & 
\multirow{3}{*}{Linear} 
& -3.42  & -3.21  & -2.85 & -1.58 & -2.07 & -1.74& -2.31 &  -2.47&-1.52  &-1.68 & -2.29  \\ &
& -3.42  & -3.17  & -2.89 & -1.59 & -2.03 & -1.72& -2.33 &  -2.45&-1.52  &-1.67 & -2.28  \\ &
& $\pm$0.02 & $\pm$0.03 & $\pm$0.05 & $\pm$0.03 & $\pm$0.03 & $\pm$0.02 & $\pm$0.03 & $\pm$0.02 & $\pm$0.02 & $\pm$0.02 & $\pm$0.00 \\ 
\cline{2-13} & 
\multirow{3}{*}{2 layer NN} 
& -2.82  & -2.49  & -2.25  & -1.27 &  -1.27 & -1.29 &-1.59 & -1.81  &-1.13 & -1.25  & -1.72\\ &
& -2.84  & -2.49  & -2.30  & -1.29 &  -1.42 & -1.37 &-1.67 & -1.85  &-1.20 & -1.24  & -1.77\\ &
& $\pm$0.02 & $\pm$0.04 & $\pm$0.03 & $\pm$0.02 & $\pm$0.08 & $\pm$0.12 & $\pm$0.04 & $\pm$0.03 & $\pm$0.03 & $\pm$0.02 & $\pm$0.02 \\ 
\cline{2-13} & 
\multirow{3}{*}{ConvNet}  
& -3.17  & -2.52  & -2.20  & -1.34 & -1.72  & -1.70  &  -1.85 & -2.45 & -1.33  & -1.56 & -1.99 \\ &
& -3.16  & -2.65  & -2.21  & -1.67 & -1.75  & -1.70  &  -1.90 & -3.03 & -1.93  & -1.74 & -2.17 \\ &
& $\pm$0.06 & $\pm$0.04 & $\pm$0.03 & $\pm$0.59 & $\pm$0.07 & $\pm$0.09 & $\pm$0.04 & $\pm$0.74 & $\pm$0.83 & $\pm$0.29 & $\pm$0.19 \\ 
\cline{2-13}& 
\multirow{3}{*}{MemNN} 
& -1.46 & -2.30  & -2.03  & -1.10 & -1.14  &  -0.98 & -1.52  &  -2.33  & -1.41  & -1.21 & -1.55\\ &  
& -1.98 & -2.45  & -2.06  & -1.57 & -1.49  &  -1.07 & -1.42  &  -2.67  & -1.50  & -1.57 & -1.78\\ &  
& $\pm$0.73 & $\pm$0.13 & $\pm$0.05 & $\pm$0.76 & $\pm$0.28 & $\pm$0.10 & $\pm$0.48 & $\pm$0.47 & $\pm$0.15 & $\pm$0.58 & $\pm$0.15 \\ 
\Xhline{2\arrayrulewidth}
      \multicolumn{2}{r||}{Estimated Optimal}                 & -1.00 & -0.49 & -0.83 & -0.71 & -0.85 & -0.47 & -0.47 &   -1.83    & -1.23  & -1.06 & -0.89\\ \hline
\end{tabular}
\caption{Reward of the different models on the 10
  games, with and without curriculum. Each cell contains 3 numbers:
  (top) best performing one run (middle) mean of all runs, and (bottom) standard deviation of 10 runs with different random initialization.
  The estimated-optimal row shows the estimated highest average reward possible for each game.  Note that the estimates are based on 
  simple heuristics and are not exactly optimal.}
\label{tab:result}
\end{center}
\end{table}





\bibliography{bibliography}
\bibliographystyle{iclr2016_conference}

\end{document}